\documentclass[conference]{IEEEtran}
\usepackage[utf8]{inputenc}
\IEEEoverridecommandlockouts
\usepackage{cite}
\usepackage{amsmath,amssymb,amsfonts}
\usepackage{float}
\usepackage{booktabs}
\usepackage{siunitx}
\usepackage{algorithmic}
\usepackage{graphicx}
\usepackage{textcomp}
\usepackage{xcolor}
\usepackage{tikz}
\usepackage{multirow}
\usepackage{booktabs}
\usepackage[hidelinks]{hyperref}
\usetikzlibrary{shapes.geometric, arrows.meta, fit, positioning}
\usepackage{graphicx} 
\def\BibTeX{{\rm B\kern-.05em{\sc i\kern-.025em b}\kern-.08em
    T\kern-.1667em\lower.7ex\hbox{E}\kern-.125emX}}
\begin{document}


\title{An Exploratory Evaluation of LLM-Assisted Rewriting of Moderate-Complexity Financial Sentences for DisCoCat-Based Sentiment Analysis}

\author{\IEEEauthorblockN{1\textsuperscript{st} Brian Llinas}
\IEEEauthorblockA{\textit{Department of Computer Science } \\
\textit{Old Dominion University}\\
Norfolk, VA, USA \\
bllin001@odu.edu}
\and
\IEEEauthorblockN{2\textsuperscript{nd} Nikos Chrisochoides}
\IEEEauthorblockA{\textit{Department of Computer Science} \\
\textit{Old Dominion University}\\
Norfolk, VA, USA \\
nikos@cs.odu.edu}
}

\maketitle


\begin{abstract}
Quantum Natural Language Processing (QNLP) provides a grammar-aware framework for text modeling, with Distributional Compositional Categorical (DisCoCat) offering one of its theoretically grounded formulations. However, prior work on financial sentiment analysis has highlighted practical limitations of DisCoCat, including parser sensitivity, high simulation cost, and difficulty handling longer, more complex sentences. In this paper, we explore an LLM-assisted preprocessing workflow that uses controlled rewriting strategies to compress, simplify, or decompose moderate-complexity financial sentiment sentences into more parser-compatible and circuit-efficient variants while preserving sentiment-bearing meaning. We compare multiple prompting strategies, LLMs, and filtering configurations against the low-complexity-only DisCoCat baseline of Stein et al. At the circuit level, prompt-based rewriting substantially reduces corpus-level complexity, with the strongest compression-based variants reducing average qubit count and gate count by more than 70\% relative to the raw moderate-complexity subset. Across repeated training runs, GPT-4.1-mini + Prompt~B achieves the highest observed mean accuracy, reaching $0.550 \pm 0.035$ compared with $0.521 \pm 0.050$ for the baseline. We further find that larger training splits do not necessarily yield better downstream performance; across the evaluated configurations, training-split size had a moderately negative association with accuracy (Pearson $r=-0.446$). Taken together, these results provide exploratory evidence that LLM-assisted rewriting can make some moderate-complexity inputs usable within the evaluated DisCoCat configuration, while identifying prompt design, filtering, and circuit-aware preprocessing as important considerations in the broader effort toward more scalable and utility-oriented QNLP for financial sentiment analysis.
\end{abstract}



\section{Introduction}
In financial analysis, predicting market behavior remains a central objective. Traditional quantitative models often rely on historical prices, trading volume, and technical indicators to identify market trends and price movements \cite{deng_combining_2011}. However, these models may not fully capture contextual and behavioral signals arising from external events, financial news, social media discussions, online comments, investor emotions, and the frequency and sentiment of public information that can shape stock price movements \cite{de_oliveira_carosia_investment_2021,deng_combining_2011}. Sentiment analysis offers a complementary approach by extracting the polarity and tone of financial news and social media discourse, helping analysts better understand how information environments influence investor reactions and subsequent stock price movements \cite{ao_sentiment_2018,ko_lstm-based_2021}. Within this context, natural language processing (NLP) has become a key tool for deriving quantitative representations from financial texts, including investor and market sentiment, emotional intensity, uncertainty, and domain-specific contextual cues embedded in news, corporate disclosures, earnings calls, and social media \cite{patwardhan_transformers_2023,mishra_review_2025,du_natural_2025,todd_text-based_2024,mishev_evaluation_2020}.

As NLP becomes increasingly important for financial sentiment analysis, modern transformer-based models provide a powerful reference point. These models can generate fluent human-like text and adapt to new tasks with limited examples, making them useful for extracting and transforming information from domain-specific text \cite{nath_new_2022}. However, these capabilities are closely tied to model scale. As language models grow larger, their training requires substantial memory, distributed GPU infrastructure, and sophisticated parallelization strategies \cite{narayanan_efficient_2021}. Thus, the challenge is not only whether NLP can extract useful signals from financial text, but also whether advanced language-processing methods can remain accessible, repeatable, and computationally feasible in domain-specific research settings.

This tension motivates the exploration of complementary paradigms for language processing. Quantum computing offers one such direction. Quantum machine-learning models can, in some cases, require less training data to achieve good generalization \cite{caro_generalization_2022}, while Quantum Natural Language Processing (QNLP) provides a framework in which grammatical structure can be incorporated directly into language representations \cite{guarasci_quantum_2022,meichanetzidis_quantum_2021,lorenz_qnlp_2023}. Rather than positioning QNLP as a direct replacement for transformer-based NLP, this perspective frames it as a complementary approach with different computational and representational trade-offs for financial sentiment analysis \cite{stamatopoulos_towards_2022}.

Within QNLP, two common approaches are Quantum-enhanced Long Short-Term Memory (QLSTM) neural networks \cite{di_sipio_dawn_2022,chen_quantum_2022} and Quantum-native Distributional Compositional Categorical (DisCoCat) models \cite{martinez_multiclass_2022,lorenz_qnlp_2023}. Prior work suggests that QLSTM currently shows stronger performance than DisCoCat in financial sentiment analysis \cite{stein_applying_2023}. However, this comparison should be interpreted with caution, since the DisCoCat experiments in that study were restricted to low-complexity inputs due to practical runtime limitations. The distinction is also substantial in terms of sentence length: the low-complexity data averages 4.9 words per sentence, whereas the moderate-complexity data averages 18.4, making the latter a longer and more challenging synthetic test condition for DisCoCat-based processing. These data should not be interpreted as a direct sample of naturally occurring financial language. This leaves open whether moderate-complexity financial sentences can be incorporated into DisCoCat-based learning under current computational constraints.

Recent work on Large Language Models (LLMs) has shown that semantic compression can be used as a preprocessing strategy to reduce input complexity and improve computational feasibility without modifying the underlying model \cite{fei_extending_2024}. Inspired by this idea, we use LLMs not as end-task predictors, but as rewriting tools to compress and decompose moderate-complexity financial sentences into simpler, sentiment-preserving forms that are more suitable for lambeq-based DisCoCat training. This motivates the central question of this paper: {\bf \bf how do LLM-assisted rewriting and screening configurations affect parser validity, circuit complexity, retained training data, and observed downstream sentiment-classification accuracy for moderate-complexity financial sentences under current computational constraints?}

To address this question, we evaluate an LLM-assisted preprocessing workflow that transforms moderate-complexity financial sentences into simpler, parser-compatible, and sentiment-preserving forms for DisCoCat-based training. The main contributions of this paper are as follows:

\begin{itemize}
    \item We introduce an LLM-assisted rewriting framework for finance-oriented DisCoCat that compresses, simplifies, or decomposes moderate-complexity financial sentences into forms that are more suitable for Bobcat parsing and lambeq-based circuit construction.

    \item We quantify the effect of rewriting on both linguistic and circuit-level complexity across the corpus, measuring token length, qubit count, circuit depth, and gate count for raw and rewritten moderate-complexity sentences.

    \item We evaluate the relationship between rewritten inputs and downstream DisCoCat performance using semantic and sentiment-preservation proxies, parser validity, retained training size, accuracy, and runtime.
\end{itemize}

Empirically, LLM-assisted rewriting can reduce corpus-level circuit complexity among successfully parsed outputs, with the strongest compression-based variants reducing average qubit count and gate count by more than 70\% relative to the raw moderate-complexity subset. At the classification level, GPT-4.1-mini with Prompt~B achieves the highest observed mean accuracy among the evaluated variants, reaching $0.550 \pm 0.035$. Although this value is higher than the $0.521 \pm 0.050$ reference result reported by Stein et al.~\cite{stein_applying_2023}, the difference should be interpreted descriptively because the comparison is not a fully matched causal experiment. Larger rewritten datasets also did not necessarily produce higher accuracy. Together, these findings provide an exploratory proof of concept and an initial step toward more scalable and utility-oriented QNLP for financial sentiment analysis, while identifying trade-offs among rewriting, filtering, parser compatibility, retained data, circuit complexity, and runtime. They do not establish utility-scale deployment, a general accuracy improvement, or an optimal preprocessing strategy.

The remainder of this paper is organized as follows. Section~\ref{background} presents the background and related work. Section~\ref{methodology} describes the evaluated workflow. Section~\ref{results} reports the experimental results. Section~\ref{discussion} discusses the findings and their implications. Finally, Section~\ref{conclusion} concludes the paper.


\section{Background and Related Work}\label{background}

This section reviews the literature that motivates the present study from three complementary perspectives. First, it outlines the DisCoCat framework and the sentence-to-circuit pipeline that underpins grammar-aware QNLP. Second, it situates the problem within financial sentiment analysis, with emphasis on prior QNLP studies and the practical limitations observed for DisCoCat on more complex inputs. Third, it reviews LLM-based semantic compression and rewriting as a possible preprocessing strategy for reducing input complexity while preserving sentiment-bearing meaning. Together, these strands of work define the gap addressed in this paper: the use of LLM-guided rewriting to make moderate-complexity financial sentences more usable for DisCoCat-based learning under current computational constraints.

\subsection{DisCoCat and QNLP}

DisCoCat is one of the main frameworks used to represent linguistic meaning compositionally in QNLP. It combines distributional semantics with categorical grammar such that syntactic reductions determine how lexical meanings are composed into sentence-level meaning \cite{coecke_mathematical_2010,meichanetzidis_quantum_2021,guarasci_quantum_2022,lorenz_qnlp_2023}. In this setting, grammar is not merely a preprocessing step; it is part of the meaning-construction process itself.

Following the standard DisCoCat formulation, a sentence is represented through three stages. First, it is parsed into a grammatical expression, typically using a categorial or pregroup-based representation of syntax. Second, that grammatical structure is converted into a DisCoCat diagram, where syntactic reductions determine how word meanings interact compositionally. Third, the resulting diagram is mapped into a trainable computational object, such as a tensor-network representation or a parameterized quantum circuit, by assigning vector-space or circuit-level realizations to grammatical types and lexical items \cite{coecke_mathematical_2010,meichanetzidis_quantum_2021,lorenz_qnlp_2023}. This formulation preserves grammatical structure while linking text to trainable representations.

\begin{figure*}[t]
  \centering
  \begin{minipage}[t]{0.46\textwidth}
    \centering
    \includegraphics[width=\linewidth]{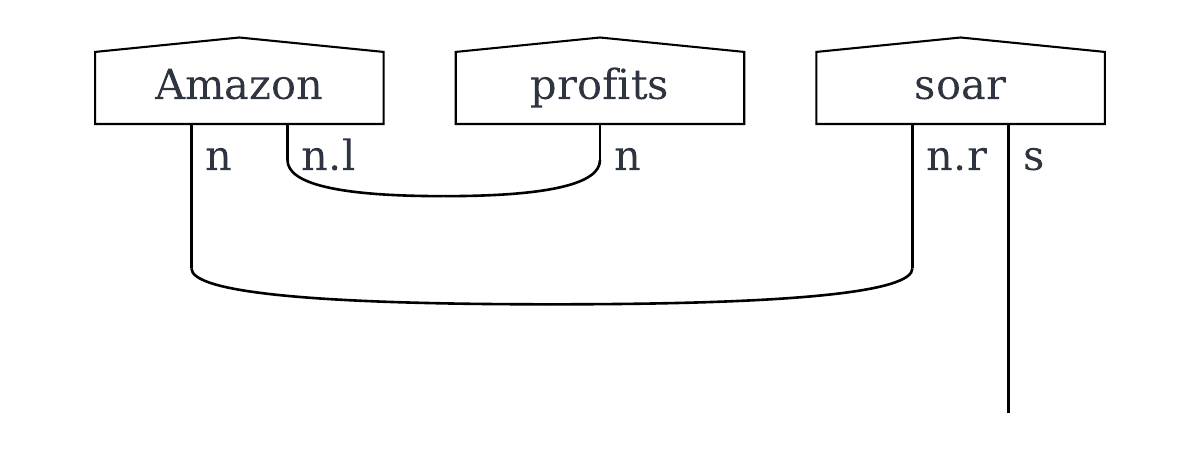}\\
    \vspace{0.4em}
    \footnotesize\textbf{(a)} Parser-derived diagram
  \end{minipage}\hfill
  \begin{minipage}[t]{0.46\textwidth}
    \centering
    \includegraphics[width=\linewidth]{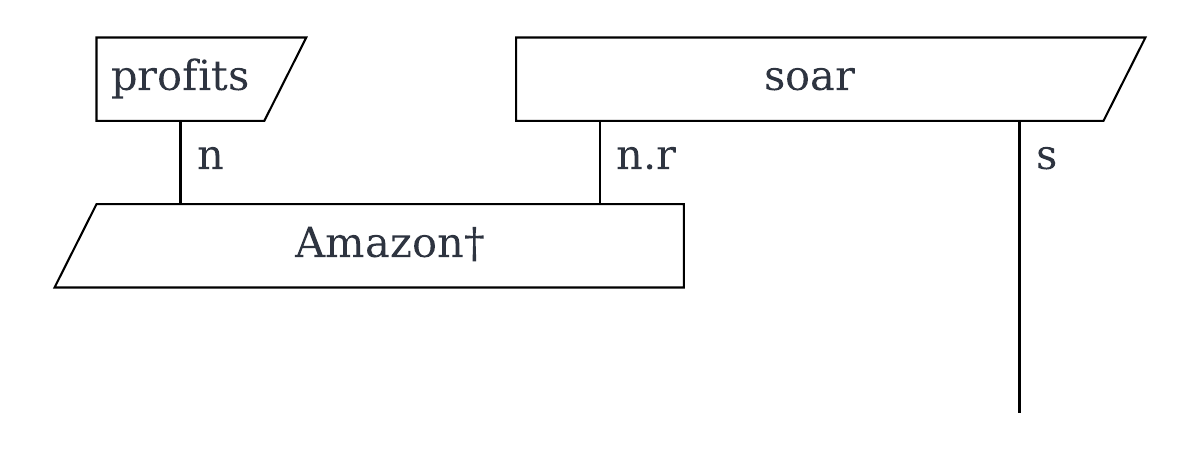}\\
    \vspace{0.4em}
    \footnotesize\textbf{(b)} Reduced diagram after cup removal
  \end{minipage}
  \caption{First two stages of the compilation process for \textit{``Amazon profits soar''}.}
  \label{fig:discocat-pipeline-diagrams}
\end{figure*}

\begin{figure*}[t]
  \centering
  \includegraphics[width=0.62\textwidth]{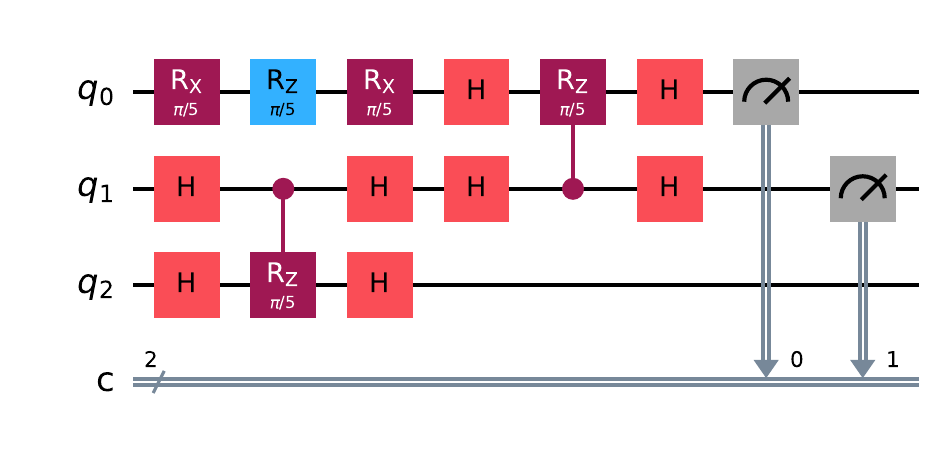}
  \caption{Final stage of the compilation process for \textit{``Amazon profits soar''}: the reduced diagram mapped into a parameterized ansatz circuit.}
  \label{fig:discocat-pipeline-circuit}
\end{figure*}

Figures~\ref{fig:discocat-pipeline-diagrams} and \ref{fig:discocat-pipeline-circuit} illustrate this pipeline for the example sentence \textit{``Amazon profits soar''}. Figure~\ref{fig:discocat-pipeline-diagrams} shows the parser-derived diagram and its reduced form after cup removal, while Figure~\ref{fig:discocat-pipeline-circuit} shows the corresponding parameterized circuit. Together, these figures make clear that the circuit ultimately used for learning is determined by the grammatical structure assigned to the sentence.

This dependency is important for the present study because sentence complexity is not only a linguistic issue but also a computational one. Since parser output determines diagram structure, and diagram structure determines the size and form of the resulting trainable representation, more complex sentences may be harder to process and may lead to greater computational burden. This issue is especially relevant in finance-oriented QNLP, where sentiment-bearing meaning often depends on modifiers, negation, and entity-rich phrasing. As a result, the applicability of DisCoCat depends not only on the framework itself, but also on whether sentence forms can be transformed into parser-compatible and trainable representations under computational constraints.

\subsection{Sentiment Analysis in Finance with QNLP}

Financial sentiment analysis is a challenging application domain because sentiment is often expressed through qualified outlook, contextual framing, and firm-specific language rather than through overtly positive or negative lexical cues \cite{ao_sentiment_2018,deng_combining_2011,de_oliveira_carosia_investment_2021,mishev_evaluation_2020,du_financial_2024,todd_text-based_2024,du_natural_2025}. As a result, financial sentiment is used not only as a classification problem in its own right, but also as a signal for downstream tasks such as forecasting, portfolio design, and market-behavior analysis.

Within QNLP, sentiment analysis has become a benchmark because it is expressive enough to require compositional reasoning while remaining manageable for current quantum and hybrid models. DisCoCat-based studies have used sentiment tasks to examine whether syntax-sensitive sentence representations can be trained end to end. Martinez and Leroy-Meline \cite{martinez_multiclass_2022} extend earlier binary settings to a multiclass sentiment task, while Ruskanda et al. \cite{ruskanda_simple_2023} show that ansatz design can affect efficiency and performance. These studies show that sentiment classification is useful not only as an application benchmark, but also as a way to study design choices within QNLP pipelines.

A related line of work evaluates quantum-enhanced recurrent or sequence-based models for sentiment-oriented tasks. Di Sipio et al. \cite{di_sipio_dawn_2022} present early QNLP experiments with a quantum-enhanced LSTM and Transformer . Chen et al. \cite{chen_quantum_2022} introduce QLSTM as a hybrid quantum-classical sequence model and report cases of faster convergence or improved accuracy . Chu et al. \cite{chu_effective_2024} further develop this line through complex-valued embeddings combined with a quantum-enhanced LSTM for sentiment analysis . These models provide a comparison point because they are sequence-oriented, whereas DisCoCat remains grammar-oriented.

The finance-specific QNLP reference point for the present work is Stein et al.\cite{stein_applying_2023}, who compare DisCoCat and QLSTM in a financial sentiment case study built from more than one thousand sentences . Their results show that QLSTM was faster to train, whereas DisCoCat remained constrained in practice. They also distinguish between low-complexity and moderate-complexity inputs, with the latter being longer on average and therefore closer to realistic financial language. This leaves open the question that motivates the present study: whether DisCoCat can be extended beyond low-complexity inputs toward sentence forms that better reflect financial language under current computational constraints.

\subsection{LLMs for Semantic Compression and Text Rewriting}

The LLM literature most relevant to this paper concerns semantic compression, simplification, and rewriting. Gilbert et al. frame semantic compression as approximate compression that preserves semantic content needed for later reconstruction or downstream use \cite{gilbert_semantic_2023}. Fei et al. \cite{fei_extending_2024} study semantic compression as a means of reducing input redundancy before downstream processing. Liskavets et al. \cite{liskavets_prompt_2025} investigate prompt compression with an emphasis on retaining information relevant to a given task or query. Across these studies, the common premise is that text can be shortened in a meaning-aware manner rather than reduced through simple truncation.

Related work on text simplification and sentence compression provides a more controlled view of this idea. Juseod-DO et al. \cite{juseon-do_instructcmp_2024} proposed InstructCMP, which shows that instruction-based LLMs can perform sentence compression under explicit length constraints. Qiang et al. \cite{qiang_redefining_2025} benchmark LLMs across lexical, syntactic, sentence, and document simplification, concluding that they outperform earlier non-LLM approaches across multiple settings. Guidroz et al. \cite{guidroz_llm-based_2025} further show that LLM-based simplification can improve comprehension and reduce cognitive load across several domains, including finance. Although these studies are not focused on QNLP, they support the use of LLMs as rewriting tools rather than only as end-task predictors.

More specifically, this rewriting behavior is typically achieved through prompting, that is, through carefully designed textual instructions that guide LLM outputs without requiring task-specific fine-tuning. The growing relevance of prompting as a mechanism for adapting LLMs to new tasks was reinforced by the strong in-context learning capabilities reported for GPT-3 \cite{ref_brown2020}, which shifted attention from parameter updates toward prompt engineering as a practical means of controlling model behavior \cite{ref_mesko2023}. Common prompting strategies include zero-shot prompting, few-shot prompting, and chain-of-thought prompting, all of which can influence how models generalize to new tasks \cite{ref_kojima2022,ref_wei2022}. In general, effective prompts specify the task, provide relevant context, and constrain the desired output format \cite{ref_giray2023}. Additional techniques such as iterative refinement and role-based prompting can further improve alignment between model outputs and task objectives \cite{ref_lin2024}. For the present study, this literature is relevant not because the LLM serves as the final classifier, but because prompt design provides the mechanism for steering the model toward controlled sentence rewriting.

For the present study, the key implication is that rewriting can be controlled. The objective here is narrower than generic summarization: rewrites must remain sentiment-bearing while also improving parser compatibility. This places the problem closer to sentence compression and simplification than to open-ended generation. Meaning-preservation metrics such as MeaningBERT \cite{beauchemin_meaningbert_2023} are useful in this context because they estimate whether rewritten sentences remain semantically close to the source. However, in finance, semantic similarity alone is insufficient, since a rewrite may remain broadly similar while still altering the sentiment signal relevant to classification. For this reason, LLM rewriting is used here as a preprocessing mechanism for transforming moderate-complexity financial sentences into forms that remain suitable for downstream DisCoCat-based learning.

\subsection{Research Gap}

Taken together, the literature suggests three observations. First, DisCoCat-based QNLP provides a grammar-aware route from text to trainable quantum representations, but its feasibility depends strongly on parser and circuit complexity \cite{meichanetzidis_quantum_2021,guarasci_quantum_2022,lorenz_qnlp_2023}. Second, financial sentiment analysis is an important application domain with structural challenges, since finance text remains domain-specific even for classical NLP systems \cite{mishev_evaluation_2020,du_financial_2024,todd_text-based_2024,du_natural_2025}. Third, LLMs are capable of semantic compression and simplification in ways that preserve much of the source meaning \cite{gilbert_semantic_2023,fei_extending_2024,juseon-do_instructcmp_2024,qiang_redefining_2025,guidroz_llm-based_2025}.

What these literatures do not yet provide is a unified treatment of rewriting as a preprocessing layer for finance-oriented DisCoCat. Existing QNLP sentiment studies focus on sentiment classification itself, ansatz design, or comparisons between QLSTM and DisCoCat \cite{martinez_multiclass_2022,ruskanda_simple_2023,stein_applying_2023}. Existing LLM compression studies focus on semantic preservation, length control, or simplification quality \cite{gilbert_semantic_2023,juseon-do_instructcmp_2024,fei_extending_2024,qiang_redefining_2025}. However, the combination of these objectives—rewriting moderate-complexity financial sentences into parser-compatible forms for Bobcat-based DisCoCat while preserving sentiment-bearing meaning—remains largely unexplored.

This gap motivates the present study. Rather than using LLMs as end-task classifiers, we use them as rewriting tools within a grammar-constrained QNLP pipeline. Rather than asking only whether financial sentiment can be classified with QNLP, we ask whether more realistic financial sentences can be transformed into forms that are usable for DisCoCat-based learning under current computational constraints. In this sense, the present study is positioned not merely as an application of LLM-based rewriting, but as a proof of concept toward more scalable and utility-oriented QNLP for financial sentiment analysis.


\section{Methodology}\label{methodology}

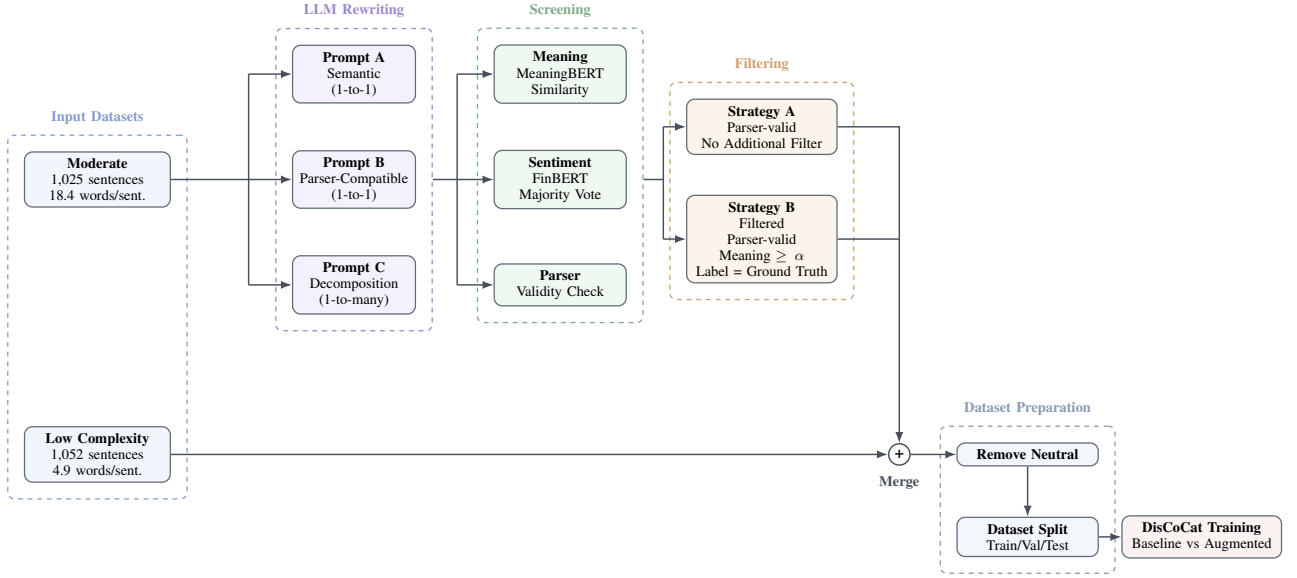
\begin{figure*}[t]
    \centering
    \resizebox{1.0\textwidth}{!}{
\definecolor{datasetfill}{RGB}{242,246,252}
\definecolor{datasetline}{RGB}{132,156,207}

\definecolor{rewritefill}{RGB}{245,241,252}
\definecolor{rewriteline}{RGB}{157,137,201}

\definecolor{screenfill}{RGB}{240,248,242}
\definecolor{screenline}{RGB}{126,170,136}

\definecolor{filterfill}{RGB}{252,244,235}
\definecolor{filterline}{RGB}{205,160,108}

\definecolor{prepfill}{RGB}{243,246,250}
\definecolor{prepline}{RGB}{137,154,178}

\definecolor{trainfill}{RGB}{250,241,241}
\definecolor{trainline}{RGB}{182,128,128}

\definecolor{textdark}{RGB}{72,78,90}
\definecolor{boxline}{RGB}{98,107,122}


\begin{tikzpicture}[
  base/.style={
    rectangle,
    rounded corners=5pt,
    draw=boxline,
    line width=0.8pt,
    align=center,
    inner sep=4pt,
    font=\small
  },
  data/.style={base, fill=datasetfill, text width=2.9cm},
  model/.style={base, fill=rewritefill, text width=2.4cm},
  eval/.style={base, fill=screenfill, text width=2.6cm},
  filter/.style={base, fill=filterfill, text width=3.0cm},
  prep/.style={base, fill=prepfill, text width=2.8cm},
  down/.style={base, fill=trainfill, text width=3.2cm},
  route/.style={draw=textdark, line width=0.9pt, sharp corners},
  arr/.style={-{Latex[length=2.4mm]}, draw=textdark, line width=0.9pt, sharp corners}
]

  \def\xdata{-1.4}
  \def\xrewrite{4.2}
  \def\xscreen{8.7}
  \def\xfilter{13.1}
  \def\xmerge{16.1}
  \def\xprep{18.9}
  \def\xtrain{22.7}


  \node[data] (moderate) at (\xdata, 0) {\textbf{Moderate}\\1,025 sentences\\18.4 words/sent.};
  \node[data] (low) at (\xdata, -6.0) {\textbf{Low Complexity}\\1,052 sentences\\4.9 words/sent.};

  \node[
    draw=datasetline,
    line width=0.8pt,
    dashed,
    rounded corners,
    fit=(low)(moderate),
    inner sep=10pt
  ] (dataset_box) {};
  \node[above=3pt of dataset_box, text=datasetline, font=\small\bfseries] {Input Datasets};

  \node[model] (prompta) at (\xrewrite, 2.3) {\textbf{Prompt A}\\Semantic\\(1-to-1)};
  \node[model] (promptb) at (\xrewrite, 0) {\textbf{Prompt B}\\Parser-Compatible\\(1-to-1)};
  \node[model] (promptc) at (\xrewrite, -2.3) {\textbf{Prompt C}\\Decomposition\\(1-to-many)};

  \node[
    draw=rewriteline,
    line width=0.8pt,
    dashed,
    rounded corners,
    fit=(prompta)(promptb)(promptc),
    inner sep=10pt
  ] (rewrite_box) {};
  \node[above=3pt of rewrite_box, text=rewriteline, font=\small\bfseries] {LLM Rewriting};

  \node[eval] (meaning) at (\xscreen, 2.3) {\textbf{Meaning}\\MeaningBERT\\Similarity};
  \node[eval] (finbert) at (\xscreen, 0) {\textbf{Sentiment}\\FinBERT\\Majority Vote};
  \node[eval] (parsercheck) at (\xscreen, -2.3) {\textbf{Parser}\\Validity Check};

  \node[
    draw=screenline,
    line width=0.8pt,
    dashed,
    rounded corners,
    fit=(meaning)(finbert)(parsercheck),
    inner sep=10pt
  ] (screen_box) {};
  \node[above=3pt of screen_box, text=screenline, font=\small\bfseries] {Screening};

  \node[filter] (strategyA) at (\xfilter, 1.15) {\textbf{Strategy A}\\Parser-valid\\No Additional Filter};
  \node[filter] (strategyB) at (\xfilter, -1.3) {\textbf{Strategy B}\\Filtered\\Parser-valid\\Meaning $\geq \alpha$\\Label = Ground Truth};

  \node[
    draw=filterline,
    line width=0.8pt,
    dashed,
    rounded corners,
    fit=(strategyA)(strategyB),
    inner sep=10pt
  ] (filter_box) {};
  \node[above=3pt of filter_box, text=filterline, font=\small\bfseries] {Filtering};

  \node[circle, draw=textdark, line width=0.9pt, fill=white, inner sep=2pt] (merge) at (\xmerge, -6.0) {\textbf{+}};
  \node[below=3pt of merge, text=textdark, font=\small\bfseries] {Merge};

  \node[prep] (remove) at (\xprep, -6.0) {\textbf{Remove Neutral}};
  \node[prep] (split) at (\xprep, -7.8) {\textbf{Dataset Split}\\Train/Val/Test};

  \node[
    draw=prepline,
    line width=0.8pt,
    dashed,
    rounded corners,
    fit=(remove)(split),
    inner sep=10pt
  ] (prep_box) {};
  \node[above=3pt of prep_box, text=prepline, font=\small\bfseries] {Dataset Preparation};

  \node[down] (training) at (\xtrain, -7.8) {\textbf{DisCoCat Training}\\Baseline vs Augmented};


  \draw[route] (moderate.east) -- (1.9, 0);
  \draw[route] (1.9, 2.3) -- (1.9, -2.3);
  \draw[arr] (1.9, 2.3) -- (prompta.west);
  \draw[arr] (1.9, 0) -- (promptb.west);
  \draw[arr] (1.9, -2.3) -- (promptc.west);

  \draw[route] (rewrite_box.east |- 0,0) -- (6.45, 0);
  \draw[route] (6.45, 2.3) -- (6.45, -2.3);
  \draw[arr] (6.45, 2.3) -- (meaning.west);
  \draw[arr] (6.45, 0) -- (finbert.west);
  \draw[arr] (6.45, -2.3) -- (parsercheck.west);

  \draw[route] (screen_box.east |- 0,0) -- (10.95, 0);
  \draw[route] (10.95, 1.15) -- (10.95, -1.3);
  \draw[arr] (10.95, 1.15) -- (strategyA.west);
  \draw[arr] (10.95, -1.3) -- (strategyB.west);

  \draw[route] (strategyA.east) -| (\xmerge, -1.3);
  \draw[route] (strategyB.east) -- (\xmerge, -1.3);
  \draw[arr] (\xmerge, -1.3) -- (merge.north);

  \draw[arr] (low.east) -- (merge.west);

  \draw[arr] (merge.east) -- (remove.west);
  \draw[arr] (remove.south) -- (split.north);

  \draw[arr] (split.east) -- (training.west);

\end{tikzpicture}
    }
    \caption{Exploratory workflow for generating, screening, filtering, and evaluating LLM-assisted rewrites of moderate-complexity synthetic financial sentences before combining them with the low-complexity reference condition.}
    \label{fig:pipeline}
\end{figure*}

This section describes the LLM-assisted preprocessing workflow used to incorporate moderate-complexity financial sentences into DisCoCat-based training. Following Stein et al. \cite{stein_applying_2023}, we use a low-complexity subset as the baseline and a moderate-complexity subset as the source for rewriting. Our main experimental comparison contrasts the low-complexity-only DisCoCat baseline of Stein et al. \cite{stein_applying_2023} with the evaluated augmented setting, in which LLM-rewritten moderate-complexity sentences are incorporated into downstream training. Figure~\ref{fig:pipeline} summarizes the workflow: we generate rewritten candidates from the moderate-complexity subset, screen them for meaning preservation, sentiment consistency, and parser validity, and merge accepted outputs with the low-complexity baseline for downstream DisCoCat training.

\subsection{Synthetic Financial Sentiment Data}

The workflow begins with two synthetic financial sentiment subsets generated through the ChatGPT-based data generation approach of Stein et al. \cite{stein_applying_2023}: a low-complexity subset and a moderate-complexity subset. The low-complexity subset serves as the baseline condition, whereas the moderate-complexity subset provides the source sentences for LLM-guided rewriting.

\begin{figure}[t]
    \centering
    \includegraphics[width=0.88\columnwidth]{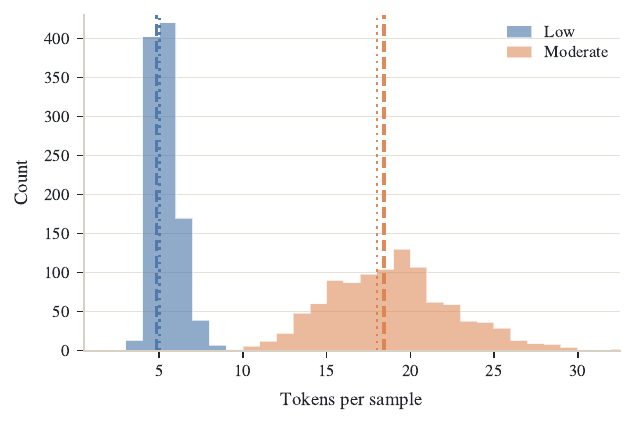}
    \caption{Token distribution for the low- and moderate-complexity subsets. Dashed lines mark means; dotted lines mark medians.}
    \label{fig:token_distribution}
\end{figure}

Table~\ref{tab:dataset_stats} and Figure~\ref{fig:token_distribution} summarize the token-length characteristics of both subsets. The low-complexity subset contains shorter inputs, ranging from 3 to 9 tokens, with a mean of 4.85 and a median of 5. In contrast, the moderate-complexity subset ranges from 10 to 32 tokens, with a mean of 18.43 and a median of 18. This distinction is important because the moderate-complexity subset more closely reflects realistic financial language, while also introducing greater challenges for DisCoCat-based processing.

\begin{table}[!t]
    \centering
    \caption{Token-length statistics for the low- and moderate-complexity subsets.}
    \label{tab:dataset_stats}
    \small
    \begin{tabular}{l
                    S[table-format=4.0]
                    S[table-format=2.0]
                    S[table-format=2.0]
                    S[table-format=2.2]
                    S[table-format=2.0]}
        \toprule
        \textbf{Subset} & {\textbf{Size}} & {\textbf{Min}} & {\textbf{Max}} & {\textbf{Mean}} & {\textbf{Median}} \\
        \midrule
        Low complexity      & 1052 & 3  & 9  & 4.85  & 5  \\
        Moderate complexity & 1025 & 10 & 32 & 18.43 & 18 \\
        \bottomrule
    \end{tabular}
\end{table}

\begin{figure}[t]
    \centering
    \includegraphics[width=0.88\columnwidth]{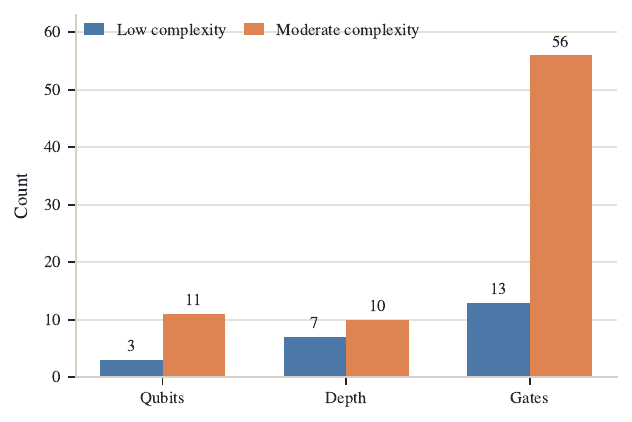}
    \caption{Illustrative comparison of circuit complexity for one low-complexity sentence (``Amazon profits soar'') and one moderate-complexity sentence (``rise of fintech has disrupted traditional banking and financial services''). The moderate-complexity example requires more qubits, greater circuit depth, and more gates than the low-complexity example.}
    \label{fig:circuit_complexity_example}
\end{figure}

This distinction also affects the complexity of the resulting DisCoCat representations. Figure~\ref{fig:circuit_complexity_example} illustrates this effect for one low-complexity sentence (``Amazon profits soar'') and one moderate-complexity sentence (``rise of fintech has disrupted traditional banking and financial services''), showing that the moderate-complexity example yields a substantially larger ansatz circuit. This difference helps motivate the rewriting stage of the proposed framework.

\subsection{LLM-Guided Rewriting Strategies}

To address this gap, we use LLM-guided rewriting strategies to transform moderate-complexity financial sentences into forms that better support downstream DisCoCat-based learning. We rely on prompting to steer LLMs toward controlled sentence rewriting rather than direct sentiment prediction. Consistent with prior work on prompt-based LLM adaptation \cite{ref_brown2020,ref_giray2023}, we design the prompts to provide clear rewriting instructions, preserve the financial context of the input sentence, and constrain the output format so that the resulting text remains compatible with downstream DisCoCat processing. In this setting, we do not seek open-ended generation; instead, we transform moderate-complexity financial sentences into simpler forms that reduce linguistic complexity while preserving sentiment-bearing meaning.

We evaluate three prompting strategies. Prompt A performs semantic compression in a one-to-one setting and produces a shorter sentence while preserving the main financial meaning and sentiment polarity. Prompt B performs parser-compatible rewriting in a one-to-one setting and adds stronger structural constraints to improve downstream parseability. Prompt C performs decomposition by mapping one original sentence into one or more shorter independent sentences when the input contains multiple separable semantic units. Table~\ref{tab:prompt_example} illustrates the three rewriting strategies on the same input sentence.

\begin{table*}[t]
\centering
\caption{Illustrative example of the three LLM-guided rewriting strategies applied to the same moderate-complexity input sentence.}
\label{tab:prompt_example}
\small
\begin{tabular}{p{4.0cm} p{3.2cm} p{3.5cm} p{5.0cm}}
\toprule
\textbf{Original sentence} & \textbf{Prompt A output} & \textbf{Prompt B output} & \textbf{Prompt C output} \\
\midrule
Stock market has been performing exceptionally well, driving up investor confidence. &
Stock market performing exceptionally well &
Stock market has been performing well. &
The stock market has been performing exceptionally well. The stock market is driving up investor confidence. \\
\bottomrule
\end{tabular}
\end{table*}

\subsection{Screening and Filtering}

After rewriting, we screen candidate outputs using three criteria: (1) semantic preservation, (2) sentiment consistency, and (3) parser validity. We measure semantic preservation with MeaningBERT \cite{beauchemin_meaningbert_2023} by comparing each rewritten output with its original sentence and obtaining a similarity score $s_i$. We evaluate sentiment consistency with FinBERT \cite{araci_finbert_2019} by comparing the predicted sentiment of the rewritten output with the original label. For Prompt~C, one original sentence may yield multiple outputs. In that case, we classify each decomposed output individually and assign the final predicted label to the original sentence by majority vote across its components. For example, if $S$ is decomposed into $s_1$, $s_2$, and $s_3$ with predicted labels $1$, $1$, and $2$, then the aggregated label for $S$ is $1$. We then compare this aggregated label with the ground-truth label. We check parser validity by attempting to parse each rewritten output with BobcatParser. We mark a candidate as parser-valid if BobcatParser converts it into a diagram without failure.

These signals define two dataset-construction strategies. Strategy~A retains all parser-valid outputs without additional filtering. Strategy~B applies stricter filtering by retaining only outputs that satisfy parser validity, sentiment consistency, and a MeaningBERT threshold, that is, $s_i \ge t$, where $t \in \{40,50,60,70,80\}$. For the final reported experiments, we use $t \ge 60$ as an exploratory operating point selected to represent a practical balance between screening score and retained training size. This threshold was not optimized using held-out data and is not treated as an optimal value.

\subsection{Dataset Construction and DisCoCat Training}

After screening, we merge the accepted rewritten outputs with the original low-complexity baseline and construct the datasets for downstream DisCoCat training. Depending on the experimental variant, we train either on the low-complexity baseline alone or on a merged dataset that combines the low-complexity baseline with rewritten moderate-complexity sentences. Let $\mathcal{D}_{L}$ denote the original low-complexity baseline and let $\mathcal{A}^{(v)}$ denote the accepted rewritten outputs for variant $v$. The resulting dataset for each variant is defined as
\begin{equation}
\mathcal{D}^{(v)} = \mathcal{D}_{L} \cup \mathcal{A}^{(v)}.
\end{equation}

In filtered variants, $\mathcal{A}^{(v)}$ contains only rewrites that satisfy the selected screening conditions, whereas in unfiltered variants it contains all available rewrites.

Since the downstream task is binary, we remove neutral instances. For Prompt~C, we assign a common group identifier to all decomposed outputs derived from the same original sentence, and we split the data at the group level to prevent information leakage across the training, validation, and test sets.

For downstream training, we parse sentences with BobcatParser and convert them into parameterized circuits using the DisCoCat configuration adopted in this study.

\subsection{Evaluation Protocol}

We evaluate the proposed framework at both the circuit level and the classification level. To characterize feasibility-related differences associated with rewriting of downstream DisCoCat processing, we measure sentence-level circuit complexity using the number of qubits, circuit depth, and number of gates. We compare these metrics between raw moderate-complexity sentences and their rewritten counterparts. For Prompt~C, which may decompose one source sentence into multiple outputs, we distinguish per-output circuit complexity and, when relevant, aggregate complexity per original source sentence.

For downstream classification, we evaluate performance at the sentence level using accuracy:
\begin{equation}
\mathrm{Accuracy} =
\frac{1}{N}
\sum_{i=1}^{N}
\mathbb{I}(\hat{y}_i = y_i),
\end{equation}
where $N$ is the number of evaluated instances, $y_i$ is the ground-truth label, $\hat{y}_i$ is the predicted label, and $\mathbb{I}(\cdot)$ is the indicator function.

When grouped outputs are present, as in Prompt~C, we also report group-level accuracy by aggregating sentence-level predictions back to the original source sentence through majority vote, with confidence-based tie-breaking when necessary.

To contextualize differences in the number of valid trainable instances produced by each preprocessing strategy, we report the data expansion ratio relative to the low-complexity baseline:
\begin{equation}
\mathrm{Expansion}^{(v)} =
\frac{n^{(v)}}{n_0},
\end{equation}
where $n^{(v)} = |\mathcal{D}^{(v)}|$ is the number of valid instances for variant $v$, and $n_0$ is the number of instances in the Stein et al. baseline.

Finally, we conduct an exploratory analysis of the association between DisCoCat training-split size and downstream accuracy across repeated runs. This training-split size is distinct from the total dataset size $n$ used in the expansion-ratio analysis. We report Pearson correlation to characterize linear association and Spearman correlation to characterize monotonic association. This analysis is descriptive rather than causal, since training size is associated with prompt, model, filtering, and retained sample composition.

\subsection{Experimental Setup}

We use \texttt{GPT-4.1-mini} and \texttt{Qwen2.5:7B} for rewriting under three prompt settings. We selected GPT-4.1-mini and Qwen2.5:7B to provide an exploratory comparison between a commercially hosted model and an open-weight model under the same rewriting and downstream DisCoCat workflow. The purpose of this comparison is to examine whether model family and deployment setting are associated with differences in semantic preservation, parser compatibility, circuit complexity, and downstream behavior; it is not intended to establish a general ranking of model quality. Because the Qwen Prompt~A and Prompt~C configurations did not complete downstream training, cross-model conclusions are restricted primarily to the Prompt~B comparison. 

For downstream DisCoCat experiments, we use the \texttt{lambeq} pipeline with the local Bobcat parser, an \texttt{IQPAnsatz}, \texttt{TketModel}, and \texttt{AerBackend} in statevector simulation mode. We split the data into 80\% training, 10\% validation, and 10\% test sets. We train with \texttt{QuantumTrainer} and \texttt{SPSAOptimizer} for 100 epochs using batch size 32, and we monitor validation accuracy during training. For comparison across variants, we report accuracy as the main evaluation metric. We implement all experiments in Python using \texttt{lambeq}, \texttt{pytket}, \texttt{qiskit}/Aer, \texttt{PyTorch}, and \texttt{transformers}. 

We run part of the experiments locally on a 14-inch MacBook Pro with an Apple M2 Pro chip, a 10-core CPU, and 16\,GB unified memory, and we run the remaining experiments on the Wahab high-performance computing environment using one NVIDIA V100 GPU, 8 CPU cores, and 192\,GB RAM per GPU node. When available, we use MPS or CUDA acceleration for rewriting, parser checks, and screening. Each reported configuration was evaluated across six repeated training runs. We report means, standard deviations, and minimum values descriptively; formal significance tests, confidence intervals, and effect-size analyses were not performed.\footnote{Code and selected experiment files are available at
\href{https://github.com/bllin001/qnlp-discocat-llms-finance-rewriting}
{https://github.com/bllin001/qnlp-discocat-llms-finance-rewriting}.}


\section{Results}\label{results}

\subsection{Effect of Rewriting on Text and Circuit Complexity}

\begin{figure*}[t]
    \centering
    \includegraphics[width=0.95\textwidth]{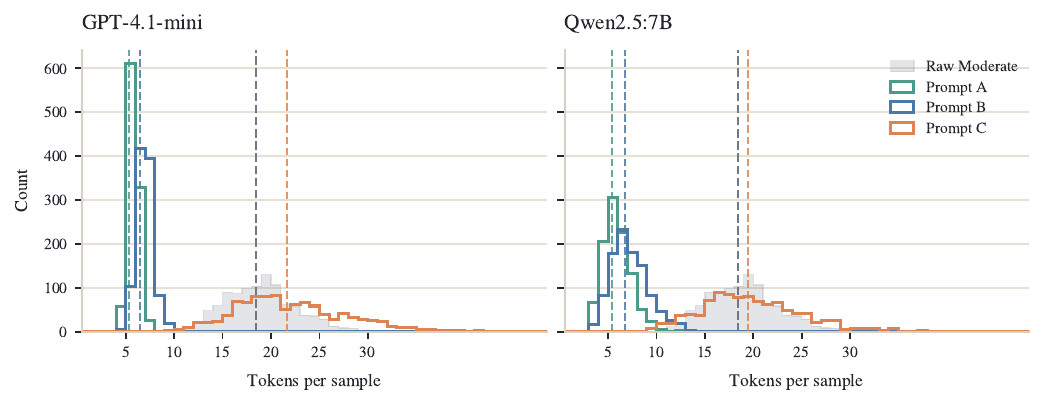}
    \caption{Token-count distributions after prompt-based rewriting of the raw moderate-complexity subset. Prompt~A and Prompt~B strongly compress sentence length, whereas Prompt~C primarily decomposes inputs into multiple shorter sentences. Dashed vertical lines indicate the mean token count for each distribution.}
    \label{fig:prompt_token_distribution}
\end{figure*}

We first examine whether prompt-based rewriting reduces the complexity of moderate-complexity financial sentences before downstream DisCoCat training. Figure~\ref{fig:prompt_token_distribution} shows the token-count distributions after rewriting for both GPT-4.1-mini and Qwen2.5:7B. Prompt~A and Prompt~B substantially compress the raw moderate-complexity inputs, shifting their distributions toward shorter lengths that are closer to the low-complexity setting. In contrast, Prompt~C behaves differently: rather than enforcing strict compression, it decomposes a single moderate-complexity sentence into multiple shorter outputs. As a result, its token-count distribution remains broader and closer to the original moderate-complexity range.

This pattern is consistent with the decomposition behavior of Prompt~C. GPT Prompt~C produces, on average, 2.72 sentences per input, with a median of 3, whereas Qwen Prompt~C produces 2.27 sentences per input, with a median of 2. More specifically, GPT decomposes most inputs into 2--3 sentences, while Qwen more often produces 2-sentence decompositions. These results indicate that Prompt~C reduces structural complexity primarily through segmentation rather than through direct token compression.

\begin{figure*}[t]
   \centering
   \includegraphics[width=0.95\textwidth]{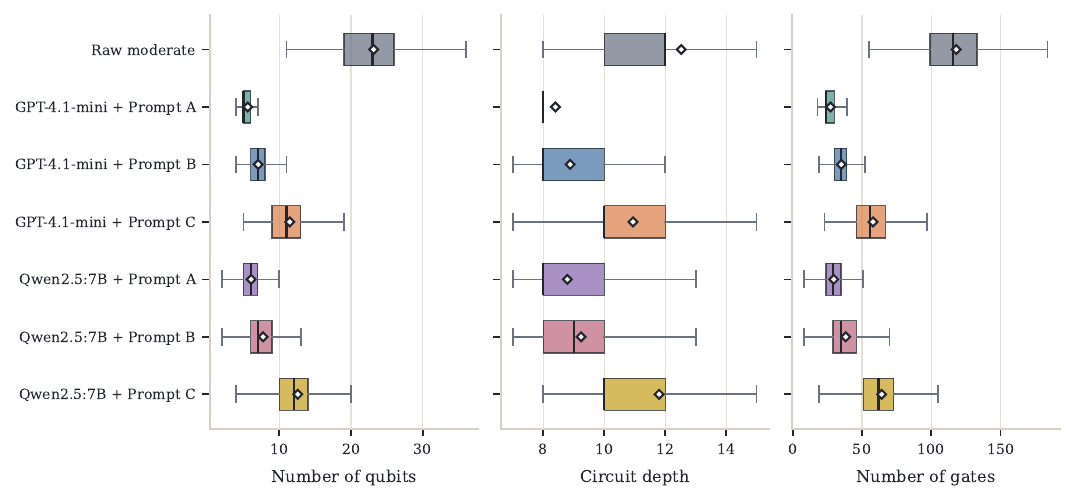}
   \caption{Corpus-level circuit complexity across rewriting variants. From left to right, the boxplots show the distributions of qubit count, circuit depth, and gate count for raw moderate-complexity sentences and their rewritten counterparts. Boxes indicate interquartile ranges, horizontal lines indicate medians, whiskers extend to the $1.5$ IQR limits, and diamonds indicate means.}
    \label{fig:corpus_circuit_complexity}
\end{figure*}

Figure~\ref{fig:corpus_circuit_complexity} extends this analysis to the circuit level at the corpus scale. Across all three metrics---qubits, circuit depth, and gates---the raw moderate-complexity sentences produce the largest circuit distributions. Prompt~A and Prompt~B generally shift these distributions toward smaller circuits, especially in terms of qubit count and total number of gates. Prompt~C also reduces circuit complexity relative to the raw moderate subset, but its distributions remain broader and typically higher than those of Prompt~A and Prompt~B. This reflects the fact that decomposition simplifies individual sentence-level circuits without necessarily minimizing the total amount of generated text.

Table~\ref{tab:corpus_circuit_complexity_summary} quantifies these corpus-level differences. Because circuit quantities are defined only for successfully parsed outputs, the reported metrics are computed over outputs that yielded valid sentence-level circuits; parser compatibility itself is analyzed separately in the next subsection. Relative to the raw moderate subset, all prompt-based variants reduce the average number of qubits, circuit depth, and gates. The strongest overall reduction is obtained by GPT-4.1-mini + Prompt~A, which lowers the average number of qubits from 23.16 to 5.60, circuit depth from 12.52 to 8.40, and gates from 118.07 to 27.29. This corresponds to reductions of 75.81\% in qubits, 32.96\% in depth, and 76.88\% in gates. Qwen2.5:7B + Prompt~A follows a similar pattern, reducing qubits by 73.75\%, depth by 29.83\%, and gates by 74.93\%.

The reductions are strongest for qubits and gates, while reductions in circuit depth are more moderate. This suggests that prompt-based rewriting is particularly effective at reducing circuit width and total gate count, whereas depth remains more sensitive to the syntactic structures produced by the parser. Prompt~C also reduces average circuit requirements relative to the raw moderate subset, but its reductions are smaller than those of Prompt~A and Prompt~B, which is consistent with its decomposition-based behavior rather than direct compression.

\begin{table*}[t]
\centering
\caption{Corpus-level circuit complexity across rewriting variants. We report mean $\pm$ standard deviation for qubits, circuit depth, and gates, together with average percentage reduction relative to the raw moderate subset. Parser compatibility is analyzed separately in Table~\ref{tab:rewriting_quality_summary}.}
\label{tab:corpus_circuit_complexity_summary}
\footnotesize
\setlength{\tabcolsep}{4pt}
\begin{tabular}{lcccccc}
\toprule
\textbf{Variant} & \textbf{Qubits} & \textbf{Depth} & \textbf{Gates} & \textbf{Qubit red. (\%)} & \textbf{Depth red. (\%)} & \textbf{Gate red. (\%)} \\
\midrule
Raw moderate
& 23.16 $\pm$ 6.07
& 12.52 $\pm$ 10.16
& 118.07 $\pm$ 31.53
& 0.00 & 0.00 & 0.00 \\

GPT-4.1-mini + Prompt A
& 5.60 $\pm$ 1.07
& 8.40 $\pm$ 0.80
& 27.29 $\pm$ 5.26
& \textbf{75.81} & \textbf{32.96} & \textbf{76.88} \\

GPT-4.1-mini + Prompt B
& 7.06 $\pm$ 1.47
& 8.88 $\pm$ 1.10
& 35.09 $\pm$ 7.46
& 69.51 & 29.10 & 70.28 \\

GPT-4.1-mini + Prompt C
& 11.46 $\pm$ 3.36
& 10.94 $\pm$ 3.46
& 57.99 $\pm$ 16.79
& 50.51 & 12.63 & 50.88 \\

Qwen2.5:7B + Prompt A
& 6.08 $\pm$ 2.53
& 8.79 $\pm$ 1.28
& 29.60 $\pm$ 12.07
& 73.75 & 29.83 & 74.93 \\

Qwen2.5:7B + Prompt B
& 7.74 $\pm$ 3.01
& 9.24 $\pm$ 1.42
& 38.29 $\pm$ 14.61
& 66.60 & 26.23 & 67.57 \\

Qwen2.5:7B + Prompt C
& 12.58 $\pm$ 4.21
& 11.80 $\pm$ 7.48
& 64.23 $\pm$ 23.28
& 45.68 & 5.79 & 45.60 \\
\bottomrule
\end{tabular}
\end{table*}

Taken together, these results show that prompt-based rewriting reduces circuit complexity through two different mechanisms. Prompt~A and Prompt~B primarily act as compression strategies, producing shorter texts that lead to substantially fewer qubits and gates. Prompt~C, in contrast, acts as a decomposition strategy: it reduces the complexity of individual sentence-level circuits but may preserve or increase the aggregate workload per original sentence. This distinction is important for interpreting downstream DisCoCat performance, since lower per-circuit complexity and lower total training cost are related but not equivalent objectives.

\subsection{Semantic Preservation, Sentiment Fidelity and Parser Compatibility}

\begin{figure}[t]
    \centering
    \includegraphics[width=\columnwidth]{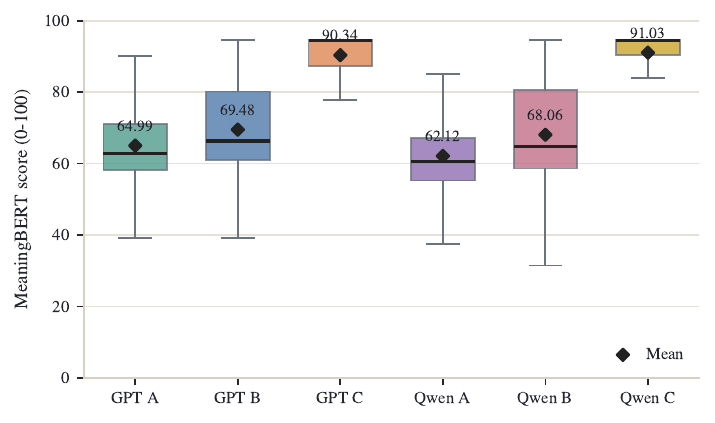}
    \caption{Distribution of MeaningBERT similarity scores for rewritten outputs. Diamonds indicate mean scores. MeaningBERT is used as an automated semantic-similarity proxy rather than a complete evaluation of semantic faithfulness.}
    \label{fig:meaningbert_summary}
\end{figure}

\begin{figure}[t]
    \centering
    \includegraphics[width=\columnwidth]{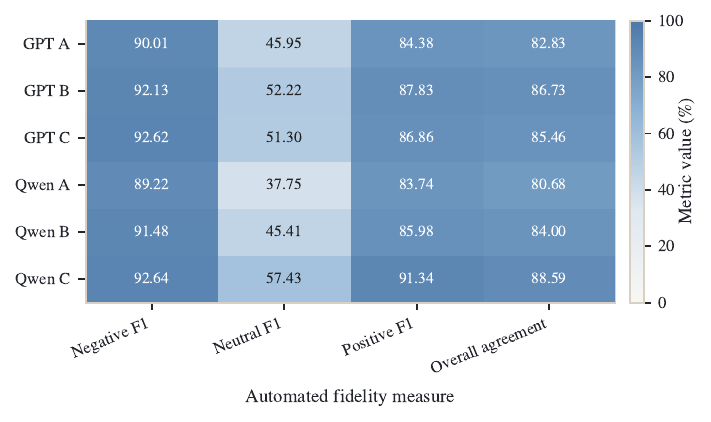}
    \caption{Automated FinBERT sentiment-fidelity diagnostics for rewritten outputs. The Negative, Neutral, and Positive columns report per-class F1 scores; Overall reports the aggregate agreement measure. Neutral sentiment is preserved less consistently than negative and positive sentiment.}
    \label{fig:finbert_summary}
\end{figure}

\begin{figure}[t]
    \centering
    \includegraphics[width=\columnwidth]{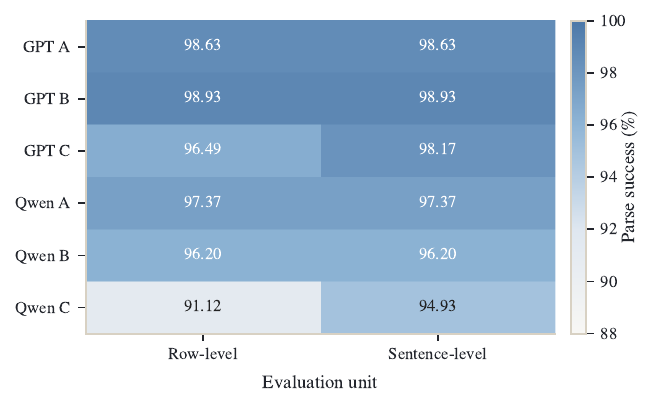}
    \caption{Bobcat parser success rates for rewritten outputs at the row and sentence levels. Values indicate the percentage of evaluation units successfully converted into valid parser diagrams.}
    \label{fig:bobcat_summary}
\end{figure}

Table~\ref{tab:rewriting_quality_summary} and Figs.~\ref{fig:meaningbert_summary}--\ref{fig:bobcat_summary} summarize rewriting quality across the three screening dimensions. Prompt~C achieves the strongest semantic preservation for both LLM families, with mean MeaningBERT scores above 90, whereas Prompt~A and Prompt~B remain in the low-to-high 60s. In sentiment fidelity, Qwen Prompt~C obtains the highest overall FinBERT agreement, although all variants preserve negative and positive sentiment more reliably than neutral sentiment. Parser compatibility remains high for most variants, with GPT Prompt~B achieving the strongest Bobcat parse success. Taken together, these results indicate that no single prompt dominates all evaluated screening criteria: Prompt~C best preserves meaning and sentiment, whereas Prompt~B best supports parseability.

\begin{table}[t]
\centering
\caption{Summary of rewriting quality across prompting variants. MB denotes mean MeaningBERT score, FB denotes overall FinBERT agreement, and BP denotes Bobcat parse success.}
\label{tab:rewriting_quality_summary}
\scriptsize
\setlength{\tabcolsep}{3pt}
\begin{tabular}{lccc}
\toprule
\textbf{Variant} & \textbf{MB} & \textbf{FB (\%)} & \textbf{BP (\%)} \\
\midrule
GPT Prompt A   & 65.0 & 83 & 98.6 \\
GPT Prompt B   & 69.5 & 87 & \textbf{98.9} \\
GPT Prompt C   & 90.3 & 85 & 96.5 \\
Qwen Prompt A  & 62.1 & 81 & 97.4 \\
Qwen Prompt B  & 68.1 & 84 & 96.2 \\
Qwen Prompt C  & \textbf{91.0} & \textbf{89} & 91.1 \\
\bottomrule
\end{tabular}
\end{table}

\begin{figure*}[t]
    \centering

    \begin{minipage}[t]{0.48\textwidth}
        \centering
        \includegraphics[width=\linewidth]{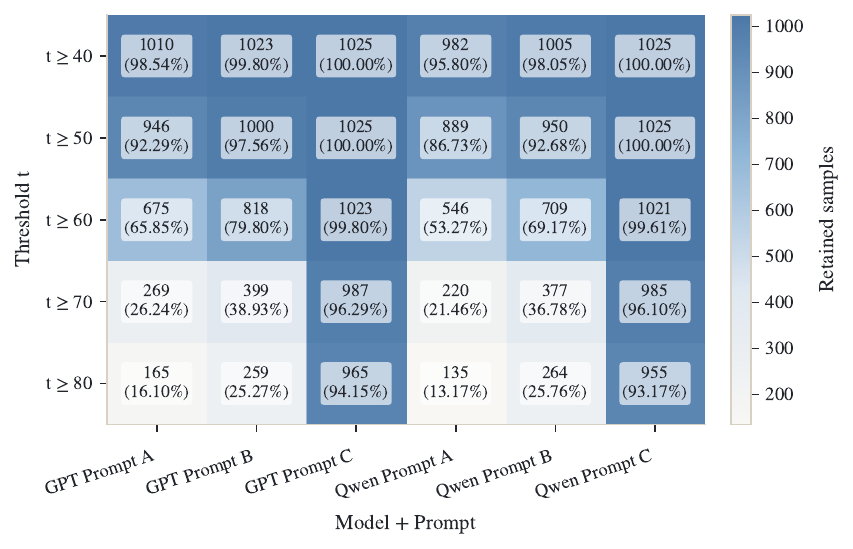}
        \par\smallskip
        \footnotesize\textbf{(a)} MeaningBERT threshold only
    \end{minipage}
    \hfill
    \begin{minipage}[t]{0.48\textwidth}
        \centering
        \includegraphics[width=\linewidth]{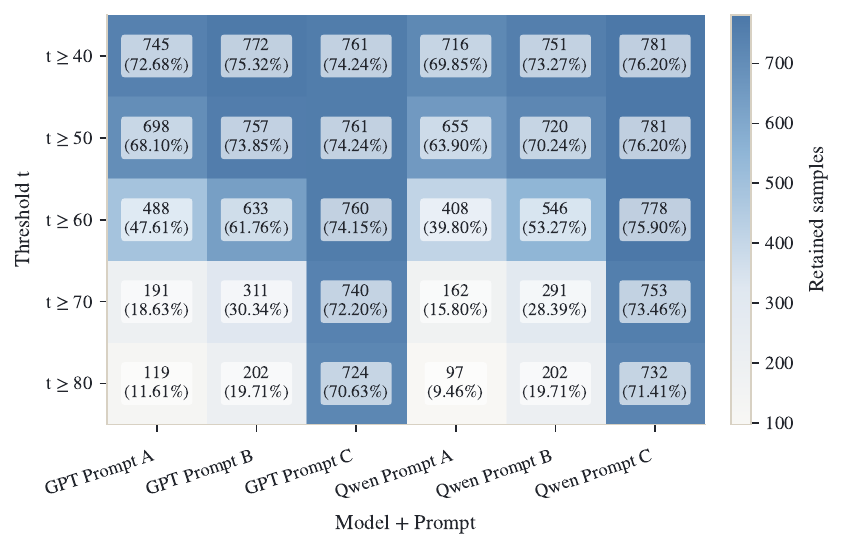}
        \par\smallskip
        \footnotesize\textbf{(b)} MeaningBERT + FinBERT agreement
    \end{minipage}

    \par\medskip

    \begin{minipage}[t]{0.48\textwidth}
        \centering
        \includegraphics[width=\linewidth]{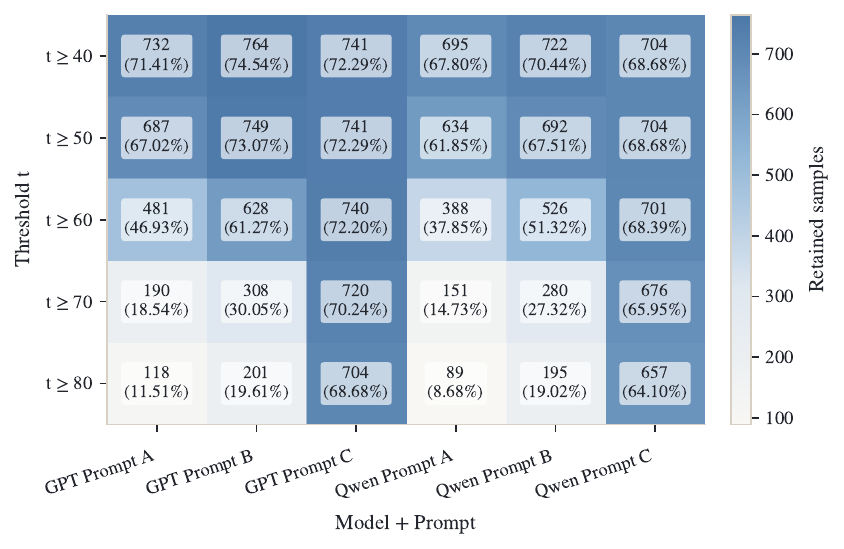}
        \par\smallskip
        \footnotesize\textbf{(c)} MeaningBERT + FinBERT + Bobcat validity
    \end{minipage}

    \caption{Training-set retention under progressively stricter screening conditions. Panel~(a) applies the MeaningBERT threshold only; panel~(b) additionally requires agreement between FinBERT predictions and the ground-truth label; and panel~(c) further requires successful Bobcat parsing. Each cell reports the retained sample count and percentage relative to the original 1,025-sentence moderate-complexity subset.}
    \label{fig:training_retention}
\end{figure*}

Figure~\ref{fig:training_retention} separates the effects of progressively stricter screening conditions. Panel~(a) applies the MeaningBERT threshold alone, panel~(b) additionally requires FinBERT agreement, and panel~(c) further requires successful Bobcat parsing. At $t \ge 60$, the full filtering condition retains 46.93\% of GPT Prompt~A, 61.27\% of GPT Prompt~B, and 72.20\% of GPT Prompt~C. The corresponding Qwen configurations retain 37.85\%, 51.32\%, and 68.39\%, respectively. We therefore use $t \ge 60$ as an exploratory operating point representing a practical balance between screening strictness and retained training size; this threshold was not optimized using held-out data.

\subsection{DisCoCat Classification Performance}

Table~\ref{tab:discocat_accuracy_size} reports the end-to-end DisCoCat classification results across training-set variants. In addition to accuracy, the table reports the total number of valid training instances produced by each preprocessing strategy, the corresponding data expansion ratio relative to the Stein et al.~\cite{stein_applying_2023} baseline, and runtime statistics. These quantities contextualize the results because they show how the evaluated preprocessing configurations change the number of usable training instances and runtime. They should not be interpreted as evidence that rewriting itself improves sentence quality, because the configurations also differ in filtering criteria, retained sample composition, and prompt structure.

Among the evaluated configurations, GPT-4.1-mini with Prompt~B achieves the highest observed mean accuracy, reaching $0.550 \pm 0.035$, while expanding the training set to $1{,}706$ instances ($1.98\times$ the Stein et al.~reference size). GPT-4.1-mini with filtered Prompt~A obtains a similar mean accuracy of $0.549 \pm 0.026$ with $1{,}317$ instances ($1.53\times$). These configurations illustrate a trade-off between retained data, observed accuracy, and runtime rather than a general improvement attributable to rewriting. The comparison with Stein et al.~\cite{stein_applying_2023} is descriptive because the experimental conditions are not fully matched.

\begin{figure*}[t]
    \centering
    \includegraphics[width=0.92\textwidth]{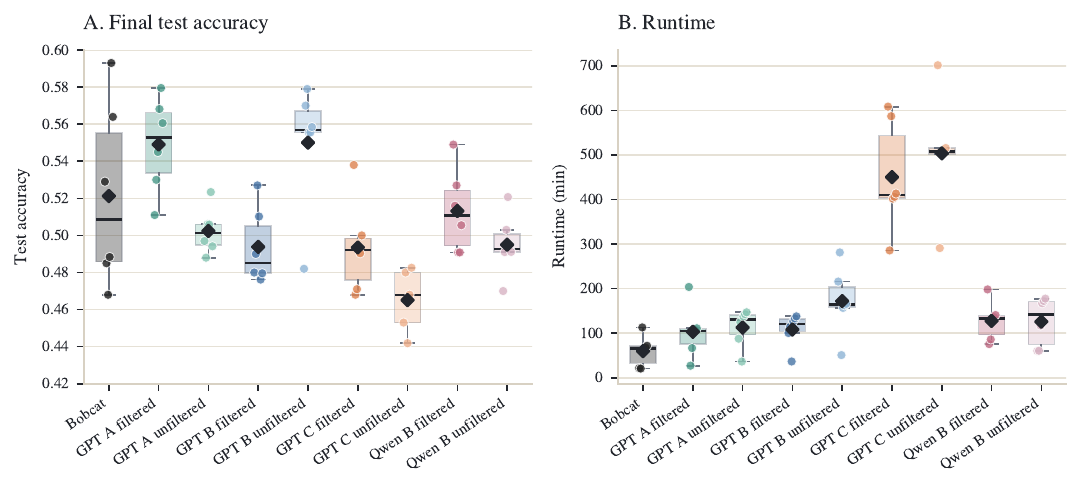}
    \caption{Run-level distribution of final test accuracy and runtime across configurations. Boxplots summarize repeated runs, and overlaid points indicate individual iterations. The figure highlights the trade-off between downstream classification performance and computational cost across prompting strategies.}
    \label{fig:run_level_performance}
\end{figure*}

Figure~\ref{fig:run_level_performance} complements Table~\ref{tab:discocat_accuracy_size} by showing the run-level distribution of final test accuracy and runtime across configurations. GPT-4.1-mini with Prompt~B and GPT-4.1-mini with filtered Prompt~A have the highest observed final test accuracies among the evaluated augmentation configurations. Filtered Prompt~A has lower mean runtime than Prompt~B, although both augmentation configurations require more runtime than the Stein et al.~reference condition. By contrast, Prompt~C variants retain more data but incur substantially higher runtime without corresponding gains in downstream accuracy.

\begin{figure*}[t]
    \centering
    \includegraphics[width=0.96\textwidth]{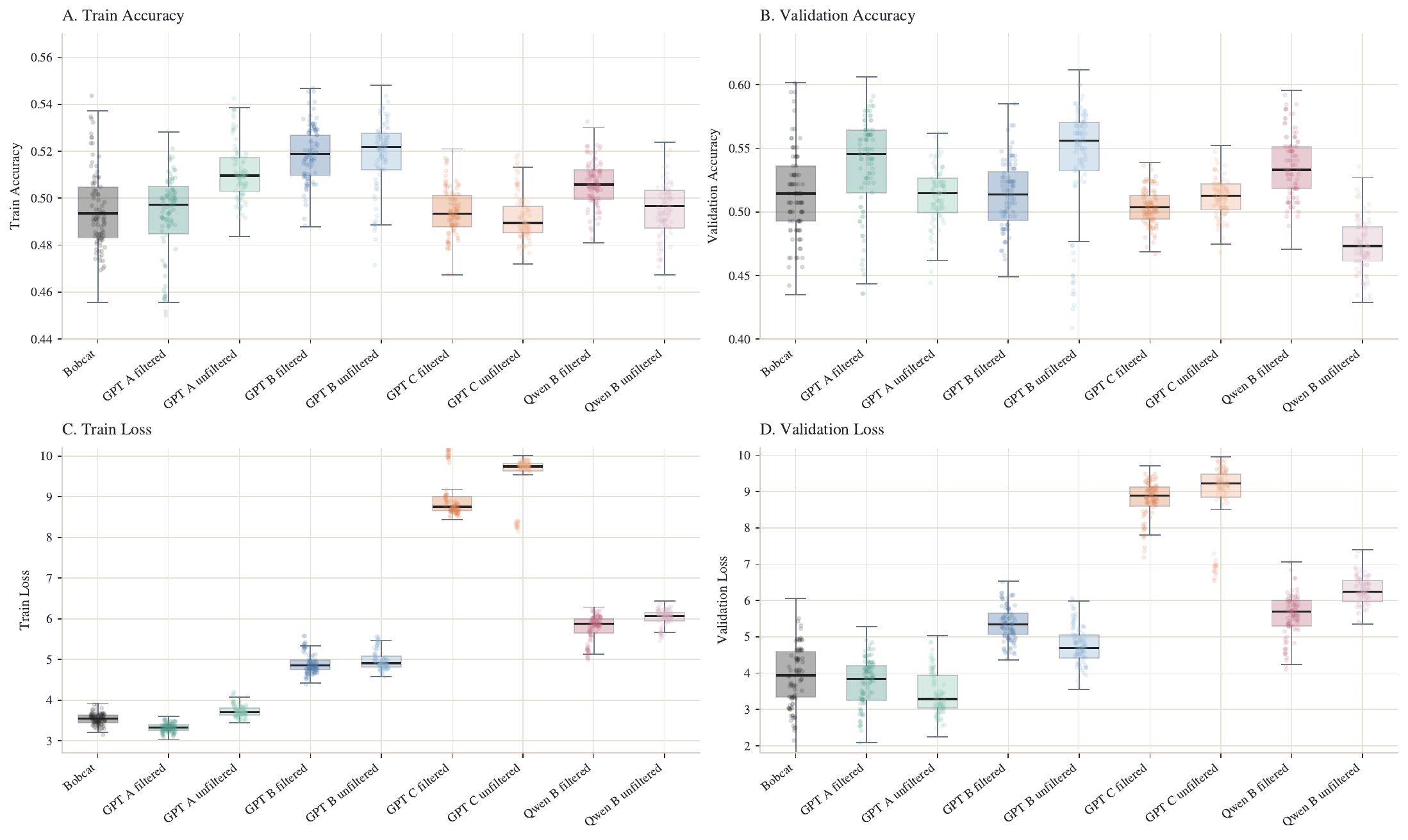}
    \caption{Epoch-level distributions of training and validation accuracy and loss across evaluated configurations. Boxplots summarize values pooled across repeated runs, and faint points show sampled epoch values. The figure describes optimization dynamics and is not a final test-accuracy comparison.}
    \label{fig:training_metric_boxplots}
\end{figure*}

Figure~\ref{fig:training_metric_boxplots} provides a complementary view of the training dynamics across configurations. The GPT-4.1-mini Prompt~B unfiltered configuration shows among the highest median training and validation accuracies, while GPT-4.1-mini Prompt~A filtered and Qwen Prompt~B filtered also show comparatively favorable validation distributions. In contrast, the Prompt~C configurations exhibit higher training and validation losses and lower or more variable accuracy distributions, consistent with their lower downstream accuracy and higher runtime in Table~\ref{tab:discocat_accuracy_size}. The effect of filtering is configuration-dependent: filtered variants are associated with higher validation medians for Prompt~A and Qwen Prompt~B, but this pattern is not uniform across all prompts. Because the boxplots pool epoch-level values across repeated runs, this figure describes optimization dynamics rather than providing an independent significance test or replacing the final test-set comparison.

However, the results also indicate that data expansion alone is not sufficient to increase downstream accuracy. The Stein et al.~reference condition contains $860$ instances, whereas GPT-4.1-mini with Prompt~B contains $1{,}706$ instances ($1.98\times$) and obtains an observed mean accuracy of $0.550 \pm 0.035$. GPT-4.1-mini with filtered Prompt~A contains $1{,}317$ instances ($1.53\times$) and obtains a similar mean accuracy of $0.549 \pm 0.026$. In contrast, the filtered and unfiltered GPT Prompt~C configurations contain $2{,}737$ ($3.18\times$) and $3{,}139$ ($3.65\times$) instances, respectively, but obtain lower mean accuracies of $0.493 \pm 0.025$ and $0.465 \pm 0.017$. Their mean runtimes also increase to $450.4$ and $503.7$ minutes, compared with $59.7$ minutes for the Stein et al.~reference condition. This pattern indicates that increasing the number of rewritten training instances does not necessarily increase accuracy and may substantially increase computational cost. The observed differences depend jointly on training-set size, prompt structure, filtering, retained sample composition, and parser compatibility.

\begin{figure*}[t]
    \centering
    \includegraphics[width=0.82\textwidth]{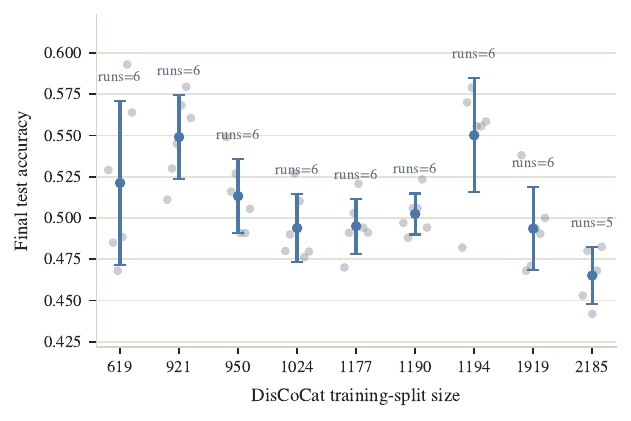}
    \caption{Observed final test accuracy across DisCoCat training-split-size groups. Gray points represent individual runs, while blue markers and error bars represent group means and standard deviations. The $n$ labels inside the figure indicate the number of repeated runs in each group, not the total dataset size.}
    \label{fig:train_size_accuracy}
\end{figure*}

Figure~\ref{fig:train_size_accuracy} summarizes accuracy across the observed DisCoCat training-split sizes. Gray points represent individual runs, whereas blue markers and error bars represent group means and standard deviations. The relationship is non-monotonic: the largest training-split groups do not produce the highest accuracy. Across the 53 runs, the association was moderately negative, with Pearson's $r=-0.446$ and Spearman's $\rho=-0.342$. Because training-split size is confounded with prompt, LLM, filtering, and retained sample composition, this pattern should be interpreted descriptively rather than causally.

Not all Qwen2.5:7B-based variants completed successfully in downstream DisCoCat training. In addition to the two Qwen Prompt~B configurations reported in Table~\ref{tab:discocat_accuracy_size}, we also attempted Qwen Prompt~A and Qwen Prompt~C under both filtered and unfiltered settings. However, these runs failed with a dimensionality-related \texttt{ValueError} during training (``setting an array element with a sequence''), indicating an inhomogeneous shape in the batched representation. We therefore exclude these failed variants from the comparative accuracy and runtime analysis. This means that the Qwen-based comparison reported here is limited to the Prompt~B setting.

\begin{table*}[t]
\caption{End-to-end comparison of DisCoCat training-set variants across repeated runs. We report the total data size $n$ after preprocessing and filtering, the data expansion ratio relative to the Stein et al. baseline ($n_0 = 860$), accuracy as mean $\pm$ standard deviation and minimum, and runtime as mean $\pm$ standard deviation and minimum (in minutes). Bold values indicate the highest observed value among the evaluated augmentation configurations.}
\label{tab:discocat_accuracy_size}
\centering
\footnotesize
\setlength{\tabcolsep}{5pt}
\begin{tabular}{lcccccc}
\toprule
\multirow{2}{*}{\textbf{Configuration}} 
& \multirow{2}{*}{\textbf{$n$}} 
& \multirow{2}{*}{\textbf{$\mathrm{Expansion}^{(v)}$}} 
& \multicolumn{2}{c}{\textbf{Accuracy}} 
& \multicolumn{2}{c}{\textbf{Runtime (min)}} \\
\cmidrule(lr){4-5}\cmidrule(lr){6-7}
& & & \textbf{Mean $\pm$ Std} & \textbf{Min} & \textbf{Mean $\pm$ Std} & \textbf{Min} \\
\midrule
Stein et al. \cite{stein_applying_2023}
& 860 & 1.00$\times$ & 0.521 $\pm$ 0.050 & 0.468 & 59.7 $\pm$ 34.8 & 20.6 \\

GPT-4.1-mini + Prompt B
& 1706 & 1.98$\times$ & \textbf{0.550 $\pm$ 0.035} & 0.482 & 172.1 $\pm$ 76.0 & 50.8 \\

GPT-4.1-mini + Prompt A + filtered
& 1317 & 1.53$\times$ & 0.549 $\pm$ 0.026 & \textbf{0.511} & \textbf{103.1 $\pm$ 58.9} & \textbf{27.0} \\

Qwen2.5:7B + Prompt B + filtered
& 1359 & 1.58$\times$ & 0.513 $\pm$ 0.023 & 0.491 & 127.6 $\pm$ 44.0 & 75.5 \\

GPT-4.1-mini + Prompt A
& 1702 & 1.98$\times$ & 0.502 $\pm$ 0.012 & 0.488 & 112.6 $\pm$ 43.1 & 36.3 \\

Qwen2.5:7B + Prompt B
& 1683 & 1.96$\times$ & 0.495 $\pm$ 0.017 & 0.470 & 125.8 $\pm$ 55.1 & 60.2 \\

GPT-4.1-mini + Prompt B + filtered
& 1465 & 1.70$\times$ & 0.494 $\pm$ 0.020 & 0.476 & 108.0 $\pm$ 37.6 & 36.5 \\

GPT-4.1-mini + Prompt C + filtered
& 2737 & 3.18$\times$ & 0.493 $\pm$ 0.025 & 0.468 & 450.4 $\pm$ 123.6 & 285.7 \\

GPT-4.1-mini + Prompt C
& 3139 & 3.65$\times$ & 0.465 $\pm$ 0.017 & 0.442 & 503.7 $\pm$ 145.3 & 291.0 \\
\bottomrule
\end{tabular}
\end{table*}


\section{Discussion}\label{discussion}

The results provide exploratory evidence that some moderate-complexity financial sentences can be transformed into forms that are usable for DisCoCat-based learning under the current parser, simulator, and training configuration. Rather than treating LLM rewriting as an end-task prediction mechanism, this study uses it as a preprocessing layer that reshapes the linguistic input before Bobcat parsing and circuit construction. In that sense, the contribution is not simply that LLMs can compress text, but that this compression can be operationalized in a way that affects the practical feasibility of downstream QNLP.

A first important finding is that rewriting quality is multi-dimensional. Results across semantic preservation, sentiment fidelity, and parser compatibility show that no single prompting strategy dominates all criteria. Prompt~C achieves the strongest meaning preservation and the highest or near-highest sentiment agreement, but these proxy scores do not establish complete preservation of propositions, entities, negation, or financial context. Prompt~B achieves the strongest Bobcat parseability, suggesting that parser-oriented prompting better supports the structural constraints of the DisCoCat pipeline. Downstream utility therefore depends on both preserving meaning and generating text that can be parsed and compiled reliably.

A second key finding is that reducing textual complexity and improving downstream performance are related, but not identical, objectives. The rewriting analysis shows that Prompt~A and Prompt~B primarily reduce sentence length, whereas Prompt~C reduces structural complexity through decomposition. These strategies therefore simplify the input in different ways. The corpus-level circuit analysis further shows that prompt-based rewriting shifts the distributions of qubits, circuit depth, and gates downward relative to the raw moderate-complexity subset, confirming that these benefits extend beyond isolated examples. Among the compression-based strategies, Prompt~A yields the strongest average reductions in circuit complexity, whereas Prompt~B provides the strongest parser compatibility and the best average downstream classification performance. In contrast, Prompt~C retains the most meaning and training data under strict filtering, yet does not achieve the best DisCoCat accuracy. Parser compatibility and downstream accuracy co-occurred in some evaluated configurations, but the present study does not isolate parser compatibility as a causal mechanism or establish that it is more important than semantic preservation.

These findings also motivate a more specialized form of decomposition. In particular, they suggest the potential value of a tone/polarity-aware decomposition strategy. A future Prompt~D could isolate distinct components of a financial sentence only when divergent or complementary sentiment-bearing units can be cleanly separated. From a circuit perspective, such a strategy could yield more compact ansatz circuits by isolating sentiment-bearing units into smaller parseable segments, potentially reducing gate count and qubit requirements while preserving compositional signal. At the same time, the present results indicate that such a strategy would need to be selective: when tone and polarity are deeply intertwined, aggressive decomposition may distort the nuanced sentiment structure required for downstream financial classification. This strategy was not evaluated in the present study and is proposed as a future extension motivated by the trade-offs observed for Prompt~C.

The filtering results reinforce this interpretation. Under increasingly strict fidelity constraints, the retained training set shrinks for all variants, but the rate of shrinkage differs substantially across prompts. Prompt~C remains much more robust at high thresholds, which makes it attractive when the primary goal is to preserve a large amount of high-fidelity training data. Nevertheless, the final experiments indicate that the most useful operating point is not the most permissive one, nor the most restrictive one. For the final experiments, we use $t \ge 60$ as an exploratory operating point that represents a practical balance between screening score and retained training size. Because this value was not selected through held-out optimization, the results do not establish it as an optimal threshold.

The final classification results show that increasing the number of rewritten training instances did not consistently increase accuracy in the evaluated configurations. GPT-4.1-mini with Prompt~B expands the training set from $860$ to $1{,}706$ instances and obtains an observed mean accuracy of $0.550 \pm 0.035$, but its mean runtime also increases from $59.7$ to $172.1$ minutes. The Prompt~C configurations expand the dataset further, to $2{,}737$ and $3{,}139$ instances, but obtain lower mean accuracies of $0.493 \pm 0.025$ and $0.465 \pm 0.017$ while requiring substantially longer runtimes. These results show that training-set expansion alone does not guarantee higher accuracy or lower cost. Because prompt type, filtering, data composition, and training-set size vary simultaneously, the study does not establish that data quality matters more than quantity. It only shows that the evaluated configurations exhibit different trade-offs among retained data, screening measures, accuracy, and runtime.

These findings position the workflow as an exploratory proof of concept rather than a complete solution. The GPT-4.1-mini + Prompt~B configuration obtained a higher observed mean accuracy than the Stein et al.~\cite{stein_applying_2023} reference result, but the comparison is not sufficient to establish superiority because the experimental conditions are not fully matched. The observed outcomes also depend on prompt design, filtering decisions, parser behavior, and the current software stack. The workflow therefore illustrates a possible way to examine the feasibility of processing longer synthetic inputs, while leaving the causal contribution of each component unresolved.

Several limitations should be noted. First, the study builds on the synthetic ChatGPT-generated financial sentiment subsets introduced by Stein et al. \cite{stein_applying_2023}. These data are useful for controlled experimentation but they do not fully capture the diversity, noise, and contextual structure of naturally occurring financial text. Second, the evaluation focuses primarily on accuracy.  Although accuracy is appropriate for comparison with the prior DisCoCat setting, it does not exhaustively characterize model behavior.

Third, although the present results now include corpus-level circuit complexity analysis, retained training size, downstream classification performance, and training runtime, the reported runtimes do not provide a stage-by-stage breakdown of rewriting, MeaningBERT screening, FinBERT screening, Bobcat parsing, circuit construction and downstream training. Decomposition-based strategies may reduce the size of individual sentence-level circuits while simultaneously increasing the number of generated outputs. Lower per-circuit complexity therefore does not automatically imply lower total pipeline cost. The training-metric boxplots also pool epoch-level values across repeated runs and should be interpreted as descriptive summaries of optimization dynamics rather than independent statistical observations.

Additional methodological limitations should also be considered. The study uses six repeated runs per configuration and reports descriptive summaries without formal significance tests, confidence intervals, or effect-size estimates. The filtering procedure combines parser validity, FinBERT agreement, and MeaningBERT thresholds, so downstream differences cannot be attributed to rewriting alone. The study does not include a classical NLP baseline, a strong non-LLM or rule-based rewriting control, or an ablation that separates the effects of rewriting from the effects of filtering and parser-based selection. MeaningBERT and FinBERT provide useful automated screening signals, but they are proxies rather than complete assessments of proposition preservation, negation, financial-entity preservation, or human-perceived faithfulness. The threshold $t \ge 60$ was used as an exploratory operating point and was not optimized using held-out validation data. The current workflow is also one-pass: outputs that fail Bobcat parsing are excluded, and parser diagnostics are not returned to the LLM for iterative repair.

Fourth, the comparison across LLM families is not fully symmetric. Although Qwen2.5:7B variants for Prompt~A and Prompt~C were attempted under both filtered and unfiltered settings, these runs failed during downstream DisCoCat training because of a dimensionality-related batching error. Consequently, the Qwen-based downstream comparison is restricted to Prompt~B, and conclusions about differences between LLM families should be interpreted cautiously. 

Finally, the conclusions remain tied to the software stack, parser behavior, simulator settings, model versions, and decoding configuration. The experiments use simulator-based execution and do not evaluate hardware execution. The study does not include human judgments of semantic faithfulness or examine overlap between language-model pretraining data and the synthetic source material. These factors may affect the generalizability and reproducibility of the results.

Future work should first establish a stronger comparative and statistical foundation for the observed results. Controlled experiments should include a classical or non-LLM rewriting baseline, a rule-based simplification baseline, and ablations that separately evaluate rewriting, filtering, parser selection, and training-set composition. Future studies should also use matched data splits, predefined threshold-selection procedures, confidence intervals, formal hypothesis tests, and effect-size estimates. In particular, the MeaningBERT threshold $t$ should be selected using held-out validation data or evaluated as a predefined hyperparameter rather than selected only from observed retention curves. These steps would help separate the effects of rewriting from those of filtering, parser selection, and the composition of the retained training sets.

Once this comparative foundation is established, a second priority is to complete and broaden the evaluation of models and semantic faithfulness. The dimensionality and batching failures affecting Qwen Prompt~A and Prompt~C should be diagnosed and resolved before drawing comparisons across model families. Subsequent experiments should evaluate newer LLM versions and expand the set of evaluated models to include additional hosted and open-weight systems across different sizes and architectures. Human evaluation of semantic faithfulness, including entity, negation, polarity, and financial-event preservation, would complement the MeaningBERT and FinBERT proxy measures. To support reproducibility, each experiment should record the model version or checkpoint, decoding settings, and relevant software-library versions.

A third priority is to characterize computational cost and scalability across the full pipeline. Future work should separately measure LLM inference, semantic screening, sentiment screening, Bobcat parsing, diagram construction, circuit compilation, and DisCoCat training. For Prompt~C, the analysis should report both per-output circuit complexity and aggregate workload per original sentence, since smaller individual circuits may still increase the total number of generated outputs. Evaluation with additional simulators, hardware backends, and updated parser versions would clarify whether the observed trade-offs depend on the current software configuration. Future work could also examine whether circuit-knitting or other distributed quantum-computing approximations can support larger DisCoCat circuits, including the weak-coupling approach proposed by Stenger et al.~\cite{stenger_scalable_2026}.

A fourth direction, motivated by the parser and circuit-construction failures observed in the current workflow, is to introduce parser-in-the-loop iterative repair. When parser failure diagnostics are available, they could be returned to the LLM together with targeted grammatical or structural constraints for a bounded retry. Each repaired output would then require re-evaluation for semantic similarity, sentiment consistency, parser validity, circuit compatibility, and downstream cost. Such an agentic repair loop should be evaluated as a separate preprocessing variant with a fixed retry budget and explicit stopping conditions, since repeated retries could increase runtime or introduce semantic drift. The goal would be to determine whether parser feedback improves successful compilation without producing unacceptable changes in sentiment-bearing meaning.

Finally, the workflow should be evaluated on naturally occurring financial corpora and across multiple financial subdomains. A specialized Prompt~D for tone- and polarity-aware decomposition remains a possible extension, but it should be evaluated as a controlled comparison rather than replace the existing prompts. Together, these directions would help determine which improvements arise from rewriting itself and which arise from filtering, parser compatibility, or changes in the composition of the retained training sets.

\section{Conclusion}\label{conclusion}

This paper presented an exploratory evaluation of LLM-assisted rewriting for incorporating moderate-complexity synthetic financial sentences into DisCoCat-based sentiment analysis. Starting from the low- and moderate-complexity datasets introduced by Stein et al.~\cite{stein_applying_2023}, we evaluated whether prompt-based rewriting could produce inputs that remain usable under the current Bobcat, lambeq, and simulator configuration.

The results show that Prompt~A and Prompt~B primarily reduce sentence length, whereas Prompt~C reduces structural complexity through decomposition. Among successfully parsed outputs, the rewriting variants reduced several circuit-complexity measures relative to the raw moderate-complexity subset. Screening results also show that the prompts exhibit different trade-offs across semantic similarity, sentiment agreement, parser validity, and retained data. These measures should be interpreted as automated proxies and not as complete evidence of meaning preservation.

For downstream classification, GPT-4.1-mini with Prompt~B achieved the highest mean accuracy among the evaluated variants, while GPT-4.1-mini with Prompt~A and filtering obtained a similar mean with lower runtime. However, the comparison with the Stein et al.~\cite{stein_applying_2023} result is descriptive rather than a matched causal evaluation, and the incomplete Qwen Prompt~A and Prompt~C runs further limit cross-model conclusions. Overall, the findings suggest that LLM-assisted rewriting can be a useful subject for further investigation of parser and circuit feasibility, but they do not establish a general accuracy improvement or an optimal preprocessing strategy.

Overall, this study provides an exploratory proof of concept and an initial step toward utility-scale QNLP for financial sentiment analysis by examining whether LLM-assisted rewriting can make some longer synthetic financial inputs usable within a reproducible DisCoCat workflow. The results do not establish utility-scale deployment, a general accuracy improvement, an optimal preprocessing strategy, or a complete solution to DisCoCat scalability. Instead, they identify trade-offs among rewriting, filtering, parser compatibility, retained data, circuit complexity, and runtime, while highlighting the preprocessing, evaluation, and scalability requirements that should be addressed through controlled future experiments.


\section*{Acknowledgment}

This research was supported in part by the Richard T. Cheng Endowment at ODU. The authors thank Min Dong of ODU ITS for assistance with HPC resources. We also thank John P. T. Stenger of the U.S. Naval Research Laboratory for his support. We thank the reviewers for their constructive feedback; due to time and space constraints, only some suggestions are reflected in this version, with additional revisions planned for a journal version. The authors acknowledge the use of Google's Gemini AI during the preparation of this manuscript. In accordance with IEEE policy, we note that this generative AI system was used strictly as an advanced copyediting and formatting assistant. All scientific concepts, experimental data, algorithms, and core analytical arguments remain original human-authored work. This work was performed using computational facilities at ODU enabled by grants from the National Science Foundation (MRI grant no. CNS-1828593), the Virginia Commonwealth Technology Research Fund, and Google Cloud Platform through ODU's Monarch Sphere initiative. Any subjective views or opinions expressed in this paper do not necessarily represent the views of the national laboratories, the NSF, or the United States Government.

\bibliographystyle{IEEEtran}
\bibliography{references}

\end{document}